\documentclass[conference]{IEEEtran}

\IEEEoverridecommandlockouts

\usepackage{times}

\usepackage[numbers]{natbib}
\usepackage{multicol}
\usepackage[bookmarks=true]{hyperref}
\usepackage{graphicx}
\usepackage{amsmath}
\usepackage{amssymb}
\usepackage{multirow}
\usepackage{array}
\usepackage{booktabs}
\usepackage[ruled,noline,noend]{algorithm2e}
\usepackage{capt-of}
\usepackage{xcolor}

\SetKwRepeat{Do}{do}{while}

\DeclareMathOperator*{\argmin}{arg\,min}

\newcommand{\manifold}{\mathcal{M}}
\newcommand{\Tspace}{\mathrm{T}}

\newcommand{\st}{\text{s.t.} \;\;\;\;\;} 
\newcommand{\stalign}{\;\;\;\;\;\;\;\;\;}
\newcommand{\derivative}{\textrm{D}}
\newcommand{\td}{\textrm{d}}  

\newcommand{\real}{\mathbb{R}}

\newcommand{\X}{X} 

\newcommand{\grad}{\text{grad}}
\newcommand{\retract}[1]{\text{R}_{#1}}

\newcommand{\transpose}{\mathsf{T}}

\newcommand{\numf}{n_{\mathrm{f}}}
\newcommand{\numg}{n_{\mathrm{g}}}
\newcommand{\numh}{n_{\mathrm{h}}}
\newcommand{\numv}{n_{\mathrm{v}}}

\newcommand{\activec}[1]{{#1}^{\mathcal{A}}}
\newcommand{\inactivec}[1]{{#1}^{\mathcal{I}}}

\newcommand{\paren}[1]{\left(#1\right)}
\newcommand{\norm}[1]{\left\lVert#1\right\rVert}
\newcommand{\bracket}[1]{\left\{#1\right\}}

\newcommand{\Xinterior}{X^{\text{Int}}}
\newcommand{\Xboundary}{X^{\partial}}

\newcommand{\rank}{\mathrm{rank}}

\begin{document}

\title{CMC-Opt: Constraint Manifold with Corners\\ for Inequality-Constrained Optimization}

\author{\IEEEauthorblockN{1\textsuperscript{st} Yetong Zhang}
\IEEEauthorblockA{\textit{College of Computing} \\
\textit{Georgia Institute of Technology}\\
Atlanta, USA \\
yetongz.gatech@gmail.com}
\and
\IEEEauthorblockN{2\textsuperscript{nd} Frank Dellaert}
\IEEEauthorblockA{\textit{College of Computing} \\
\textit{Georgia Institute of Technology}\\
Atlanta, USA \\
frank.dellaert@cc.gatech.edu}
}

\maketitle

\begin{abstract}
We introduce a manifold-based framework for addressing optimization problems with equality and inequality constraints found in robotics.
Our approach transforms the original problem into an unconstrained optimization problem directly on the constrained state space.
To achieve this, we introduce ``constraint manifolds with corners" to represent the state space satisfying mixed nonlinear equality and inequality constraints.
We further extend manifold optimization algorithms to operate on this new topological structure.
We demonstrate the power and robustness of our framework in the context of a large-scale kinodynamic planning problem, successfully generating dynamically feasible trajectories where standard methods fail.
\end{abstract}
\vspace{-0.5em}

\section{Introduction}
Modern robotic systems are inherently governed by complex hard constraints. 
They arise from diverse sources: rigid-body kinodynamic constraints~\cite{modern-robotics}, contact constraints~\cite{mastalli2020crocoddyl} (e.g., Coulomb friction, complementarity constraints\cite{Posa14ijrs_contact}), hardware limitations (e.g., joint angle, speed, and torque limits), and task-specific requirements (e.g.,  collision avoidance). While discrete-valued constraints  also exist in scenarios like footstep planning \cite{Kuindersma16ar_atlas_optimization}, they fall outside the scope of this paper, which focuses exclusively on continuous state spaces.

Standard constrained optimization methods struggle to balance cost reduction with strict constraint satisfaction in highly nonlinear robotic state spaces.
Methods that incorporate constraints into the cost function via penalties or Augmented Lagrangians \cite{Kalarishnan11icra_stomp, Zucker13ijrr_chomp, Schulman14ijrr_trajopt, Mukadam18ijrr_gpmp2, bertsekas1997nonlinear, Nocedal06book_NumericalOptimization, Vandenberghe20slide_ECNLS} must scale weights on penalty terms toward infinity to enforce exact feasibility, which can lead to numerical conditioning issues or slow convergence.
For algorithms that attempt to satisfy optimality conditions, like sequential quadratic programming (SQP) and primal-dual interior point methods \cite{cunningham10ddf_constraint, ta14_constraints, Byrd08_inexact_sqp_ecopt, betts2010practical}, devising a merit function for global convergence is a nontrivial task.

A compelling alternative to constrained optimization methods is \textit{manifold optimization}, offering lower complexity and better numerical properties~\cite{Absil07book_manifold_optimization, Boumal20book_intromanifolds, mahler2024manifold}.
Manifolds are devised to represent feasible state spaces defined by equality constraints.
Optimization is then directly conducted on these manifolds, which constitute a lower-dimensional state space.
\cite{Zhang23icra_cmopt} demonstrated the applicability of manifold optimization in solving equality-constrained nonlinear problems.

However, the standard manifold theory breaks down when introduced with inequality constraints. The feasible state spaces contain sharp ``boundaries" and ``corners" that are not locally homeomorphic to the Euclidean space, which breaks the smoothness assumptions used in manifold optimization.

In this paper, we extend the constraint manifold optimization framework~\cite{Zhang23icra_cmopt} to operate on state spaces defined by both nonlinear equality and inequality constraints.
We make the following core contributions:
\begin{itemize}
    \item We formulate \emph{constraint manifolds with corners} (CMC) to represent state spaces governed by mixed nonlinear equality and inequality constraints, and provide their local geometric parameterization.
    \item We present a general framework that transforms the standard constrained optimization problem into an unconstrained problem on the constrained state space. This is achieved by leveraging factor graphs to isolate constraint-connected components and turn them into CMCs.
    \item We extend manifold optimization techniques to operate on CMCs, adapting Riemannian gradient descent to handle sharp manifold boundaries and corners.
\end{itemize}

\begin{figure}[t]
\centerline{\includegraphics[width=0.48\textwidth]{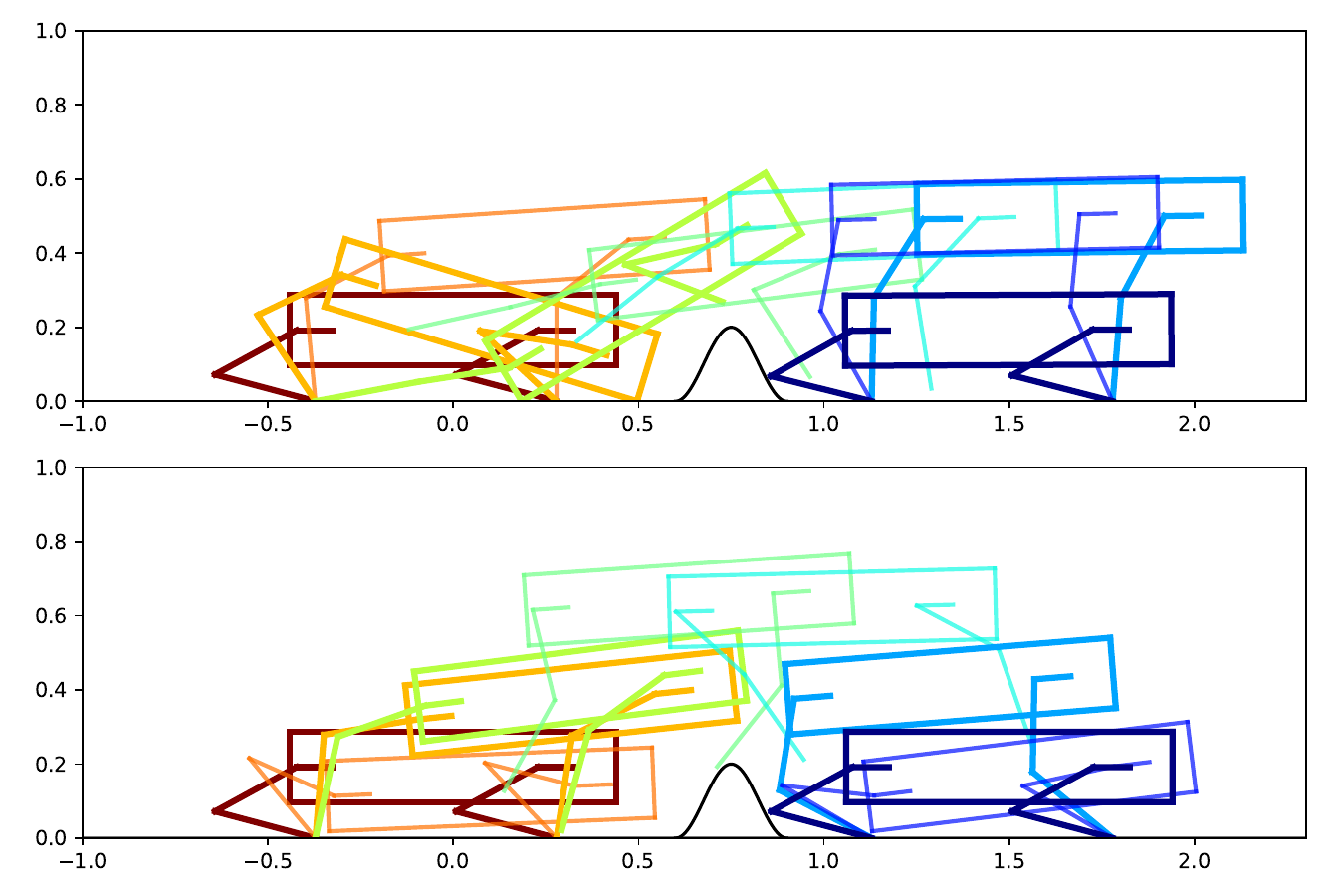}}
\caption{The optimized quadruped jumping trajectory with the penalty method (top) and the CMC-Opt method (bottom). The trajectory produced by the CMC-Opt method exhibits a smooth and natural jumping motion. In contrast, the trajectory generated by the penalty method renders dynamically infeasible due to significant "collocation leakage" where physics are slightly violated to minimize torque costs.}
\label{fig:quadruped_forward_jump}
\vspace{-1em}
\end{figure}

\newpage
\section{Problem Formulation}
We consider a general nonlinear constrained optimization problem \eqref{eqn:constrained_opt_problem}, and model it using a factor graph framework \cite{Kschischang01it_factor_graphs, Dellaert17fnt_FactorGraphs}. The continuous state space is defined by variable nodes $\X=\{x_1,\cdots,x_{\numv}\}\in\real^N$.
The three types of factor nodes encode smooth objective functions $f_j (X_{I^{\mathrm{f}}_j})$, nonlinear equality constraints $h_k(X_{I^{\mathrm{h}}_k})=0$, and nonlinear inequality constraints $g_\ell(X_{I^{\mathrm{g}}_\ell})\geq 0 $, where $I^{\mathrm{f}}_j$,  $I^{\mathrm{h}}_k$, $I^{\mathrm{g}}_\ell$ represent the multi-index sets indicating the variables involved in the cost function or constraint expression.
Traditional solvers attempt to solve \eqref{eqn:constrained_opt_problem} directly in $\real^{N}$. 
\begin{subequations}
\begin{align}
  \X^\star =  &\argmin\limits_{ \X \in \real^{N}} \sum\limits_{j=1}^{\numf} f_j (X_{I^{\mathrm{f}}_j}) \label{eqn:constrained_opt_costs}\\
    &\st h_k(X_{I^{\mathrm{h}}_k})=0 \;\;\;\;\;\;\text{for}\;\;k=1,\ldots,\numh \label{eqn:constrained_opt_constraints}\\
    &\stalign\; g_\ell(X_{I^{\mathrm{g}}_\ell})\geq 0 \;\;\;\;\;\;\text{for}\;\;\ell=1,\ldots,\numg \label{eqn:constrained_opt_i_constraints}
\end{align}\label{eqn:constrained_opt_problem}
\end{subequations}

\begin{figure}[h]
\centerline{\includegraphics[width=0.42\textwidth]{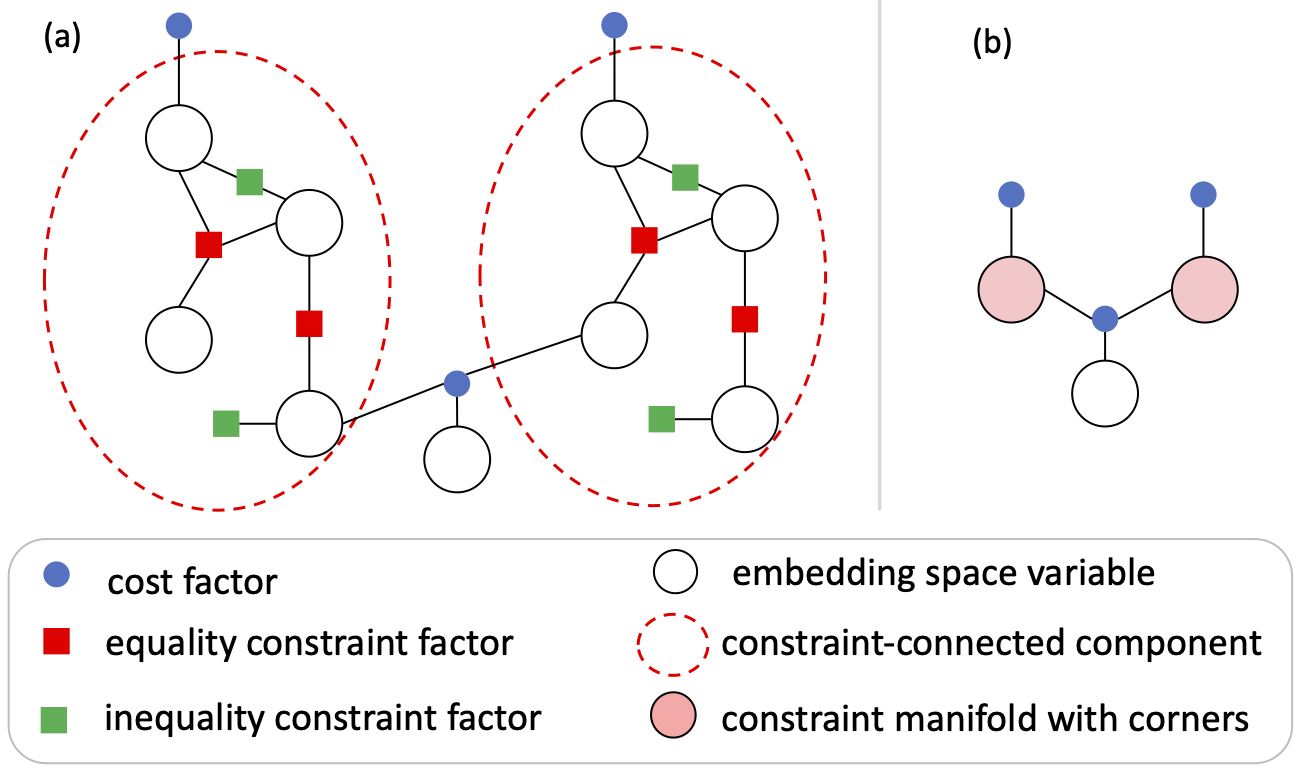}}
\vspace{-0.8em}
\caption{(a) Factor graph representation of a constrained optimization problem. (b) Factor graph representation of the transformed unconstrained optimization problem on constraint manifolds with corners.}
\label{fig:fg_constraint_groups}
\end{figure}

Our approach aims to fold the constraints \eqref{eqn:constrained_opt_constraints} and \eqref{eqn:constrained_opt_i_constraints} into the geometry of the search space. With the factor graph representation, we can easily identify sets of variables connected by local constraints that involve only variables within the same set, as demonstrated in Figure~\ref{fig:fg_constraint_groups}. Our objective is to formulate the feasible state-space of such \emph{constraint-connected components} and formulate optimization algorithms on them. 

\section{Constraint Manifold with Corners}
We consider the state space constrained by both equality and inequality constraints, embedded in $\real^{N}$.
Let $h:\real^{N}\rightarrow\real^{\numh}$ and $g:\real^{N}\rightarrow\real^{\numg}$ be functions representing the complete set of equality and inequality constraints, respectively,
the constrained state space $\manifold$ is then expressed as \eqref{eqn:cmc_def}.
\begin{align}
    \manifold=&\left\{X\in\real^{N}: h(X)=0,\; g(X)\geq 0 \right\}
    \label{eqn:cmc_def}
\end{align}

The constrained state space $\mathcal{M}$ exhibits ``boundaries" and ``corners"
\footnote{In differential geometry, boundaries typically refer to the space where only one inequality constraint is active; corners refer to the state space where one or more inequality constraints are active.}
where one or more inequality constraints are active, thus termed a \emph{constraint manifold with corners} (CMC). 
At a point $X\in\manifold$, we classify the inequality constraints into active constraints $\activec{g}=g_{I^{\mathcal{A}}_{X}}$~\eqref{eqn:active_i_constraints} and inactive constraints $\inactivec{g}=g_{I^{\mathcal{I}}_{X}}$~\eqref{eqn:inactive_i_constraints} based on the evaluation of the constraint function $g(X)$.
$X$ is termed an \emph{interior point} $\Xinterior$ if no inequality constraint is active; otherwise, it is termed a \emph{corner point} $\Xboundary$.
\begin{align}
    I^{\mathcal{A}}_{X} =&\bracket{\ell\in \bracket{1,\ldots,\numg}: g_\ell(X_{I^{\mathrm{g}}_\ell}) = 0}\label{eqn:active_i_constraints}\\
    I^{\mathcal{I}}_{X}=& \bracket{\ell\in \bracket{1,\ldots,\numg}: g_\ell(X_{I^{\mathrm{g}}_\ell})>0}\label{eqn:inactive_i_constraints}
\end{align}
We assume that at any point $X\in\manifold$, all the active constraint functions $h(X)$, $\activec{g}(X)$ and the cost functions $f(X)$ are continuously differentiable.
In addition, the Jacobian matrix of the equality and the active inequality constraint functions are always full rank as in \eqref{eqn:cmc_full_rank}.
\begin{align}
    \forall X\in\manifold:\;\; \rank \paren{\begin{bmatrix}
    \nabla h(X)^\transpose \\ \nabla \activec{g}(X)^\transpose
    \end{bmatrix}}
    =& \numh + \left|I^{\mathcal{A}}_{X}\right| \label{eqn:cmc_full_rank}
\end{align}

At an interior point $\Xinterior$, the constrained state space $\mathcal{M}$ is locally identical to a constraint manifold $\overline{\manifold}$ solely defined by equality constraints \eqref{eqn:cm_def}~\cite{Zhang23icra_cmopt}.
Hence, $\mathcal{M}$ is locally homeomorphic to $\mathbb{R}^n$ at $\Xinterior$, where $n=N-\numh$.
Thus, around $\Xinterior$, $\mathcal{M}$ can be locally parameterized with $\theta\in\mathbb{R}^n$.
\begin{align}
    \overline{\manifold} = \bracket{X\in\real^{N}: h(X)=0} \label{eqn:cm_def}
\end{align}

At a corner point $\Xboundary$, moving in any direction $v$ such that $\nabla g_\ell(\Xboundary)^\transpose v < 0$ will violate the inequality constraint, therefore going outside $\manifold$.
The constrained state space $\manifold$ is homeomorphic to the model space $H_m^n$~\eqref{eqn:corner_space}, resembling a corner~\cite{lee12book_smooth_manifolds, joyce09manifolds_corners}, where $m=n-\dim I^\mathcal{A}_{\Xboundary}$.
Therefore, around $\Xboundary$, $\manifold$ can be locally parameterized with $\theta\in H_m^n$.
\begin{align}
    H_m^n=&\{\theta\in\real^{n}:\theta_{1},\ldots,\theta_{n-m}\geq 0\} \label{eqn:corner_space}
\end{align}

We further define the connected subset of $\manifold$ where a subset $I$ of inequality constraints remain active as a ``corner", which is locally identical to the constraint manifold $\overline{\manifold}^{I}$ in \eqref{eqn:corner_def}.
\begin{align}
    \overline{\manifold}^{I} =& \bracket{X\in\real^{N}: h(X)=0, g_{I}(X)=0} \label{eqn:corner_def}
\end{align}

\begin{figure}[b]
\centerline{\includegraphics[width=0.45\textwidth]{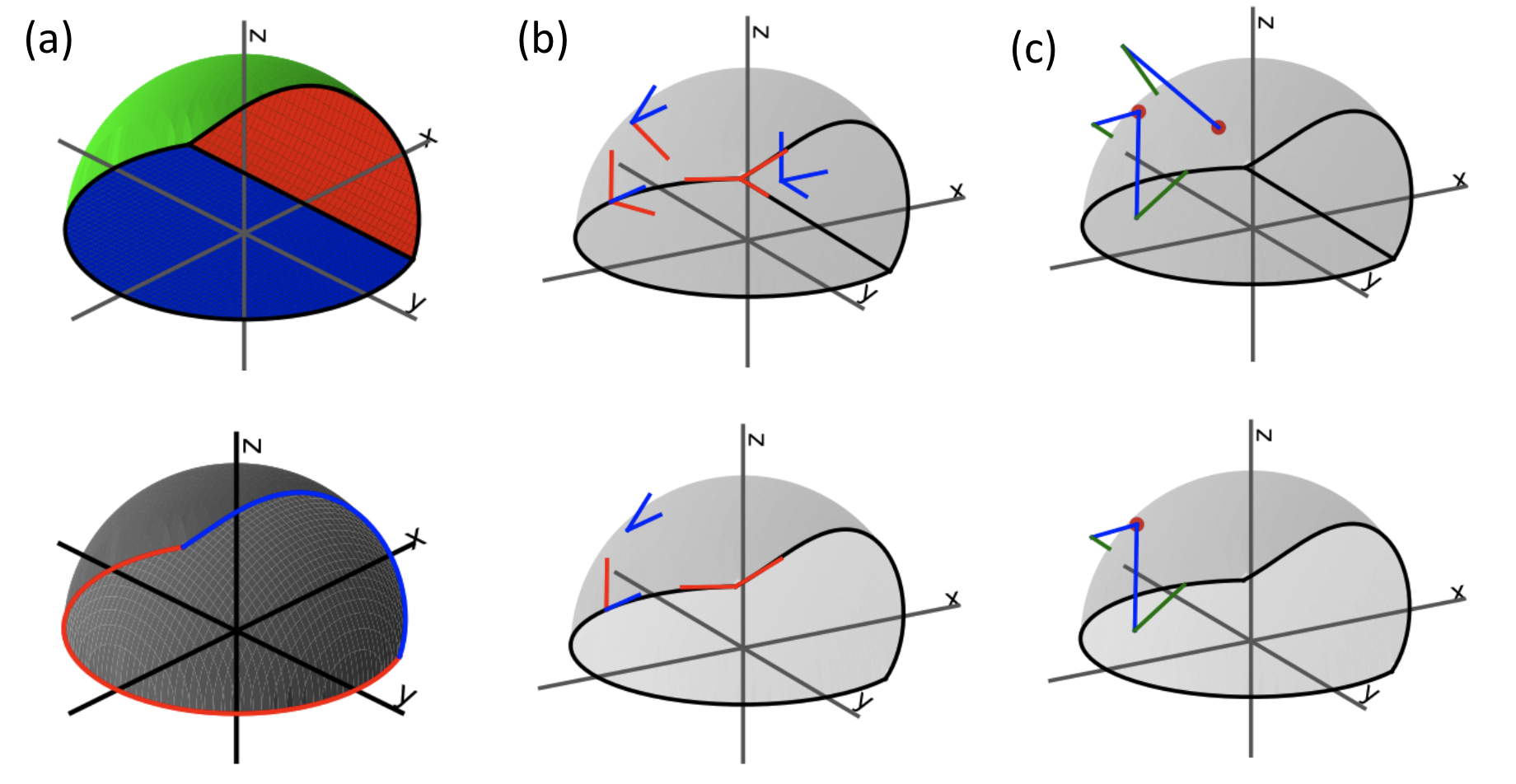}}
\vspace{-0.8em}
\caption{(a) Example constraint manifolds with corners. Top is defined by $1-x^2-y^2-z^2\geq 0$, $z-x^3+0.1 \geq 0$, $z\geq 0$; bottom is defined by $1-x^2-y^2-z^2= 0$, $z-x^3+0.1 \geq 0$, $z\geq 0$. The ``corners'' are marked with colors. (b) The tangent spaces at various points on the CMCs, represent as a set of basis vectors. Blue basis vectors indicate the freedom to move in both directions, while the red ones are limited to only move in one direction. (c) Examples of retraction operation (green) given a tangent vector (blue).}
\label{fig:tspace_retraction}
\end{figure}


\newpage
\subsection{Tangent Space}
In order to run gradient-based optimization on CMCs, we need to compute the differential of functions on CMCs, which requires the definition of a ``tangent space" that captures the local smoothness structure at any point on the CMC.

We define the tangent space of $\manifold$ following the tangent space definition of a regular manifold~\cite{Boumal20book_intromanifolds}.
The tangent space of $\manifold$ at a point $X\in\manifold$ is the set of velocities of all smooth curves $C_{X}$ on $\manifold$ starting at $X$~\eqref{eqn:smooth_curves_def}, which can be derived as the set of vectors satisfying the linearized equality and active inequality constraints (\ref{eqn:imanifold_tspace}).
Notably, the tangent space is a linear space at interior points, and a convex cone for corner points, as depicted in Figure~\ref{fig:tspace_retraction}b.
\begin{align}
    C_{X} =& \left\{
    c: [0,1) \rightarrow \manifold \text {\;is smooth}, 
    c(0)=X
    \right\}\label{eqn:smooth_curves_def} \\
\begin{split}
    \Tspace_{X}\manifold =& \left\{
    c'(0): c\in C_{X}
    \right\} \\
    =& \left\{
    v\in\real^{N}: \nabla h(X)^\transpose v=0,
    \nabla \activec{g} (X)^\transpose v \geq 0
    \right\} 
\end{split}\label{eqn:imanifold_tspace}
\end{align}
The tangent space $\Tspace_{X}\manifold$ is a subset of the tangent space of the constraint manifold $\overline{\manifold}$ defined by only the equality constraints~\cite{Zhang23icra_cmopt} as \eqref{eqn:cmc_basis}. Therefore, it can be parameterized with a set of basis vectors written in matrix form as $B$. Notice that for a specially devised basis, we can have $\Omega = H_m^n$.
\begin{align}
    \begin{split}
    \Tspace_{X}\manifold =& \bracket{v\in\Tspace_{X}\overline{\manifold}:  \nabla\activec{g}(X)^\transpose v\geq 0} \\
    =& \bracket{B\cdot \theta: \theta\in\Omega}
    \end{split}\label{eqn:cmc_basis}\\
    \Omega=&\bracket{\theta\in\real^{n}: \nabla\activec{g}(X)^\transpose B\theta \geq 0} \label{eqn:cmc_basis_param_def}
\end{align}

\subsection{Differential of Function on CMCs}
We now derive the differential of functions on CMCs.
Let's consider a real-valued function defined on a CMC $\mathcal{F}:\mathcal{M}\rightarrow \mathbb{R}$; the differential of the function $\mathcal{F}$ at $X\in\mathcal{M}$ is represented as a map $\mathrm{d}\mathcal{F}: \Tspace_{X}\mathcal{M} \rightarrow \mathbb{R}$, indicating the rate of change of the function value when moving from $X$ in the direction of a tangent vector $v\in\Tspace_{X}\manifold$, as defined in Equation \eqref{eqn:CMC_differential_def}.
\begin{align}
    \derivative \mathcal{F}(X) [v] =& \frac{\td}{\td t}\mathcal{F}\paren{\retract{\X}\paren{t\cdot v}
    }\bigg|_{t=0} \label{eqn:CMC_differential_def}
\end{align}
Using basis $B$ of the tangent space, the differential can be elegantly expressed in matrix form ~\eqref{eqn:CMC_differential}. Here, $f:\mathbb{R}^{N}\rightarrow \mathbb{R}$ denotes a smooth extension of $\mathcal{F}$ to the ambient space, which corresponds to the cost function in the original constrained optimization problem. 
The differential of $\mathcal{F}$ in the direction of any tangent vector $v = B\theta \in\Tspace_{X}\manifold$ adheres strictly to the property (\ref{eqn:CMC_differential_property}).
\begin{align}
    \nabla \mathcal{F}(X) =& B^\transpose \nabla f(X) \label{eqn:CMC_differential}\\
    \derivative \mathcal{F}(X) [B\theta] =& \nabla \mathcal{F}(X)^\transpose \cdot \theta  \label{eqn:CMC_differential_property}
\end{align}
The gradient of function $\mathcal{F}$ seeks a feasible direction in the tangent space that can most rapidly decrease the function value.
It can be computed as \eqref{eqn:cmc_gradient}.
\begin{align}
    \begin{split}
        &\theta^\star = \argmin\limits_{\theta\in\Omega}\norm{\nabla \mathcal{F}(X) - \theta}^2\\
        &\grad \mathcal{F}(X) = B\cdot\theta^\star
    \end{split}\label{eqn:cmc_gradient}
\end{align}

\subsection{Retraction}
We formulate the retraction operation to ensure staying on the CMC after each update step in optimization.
Similar to the definition of retraction operation of a standard manifold~\cite{Boumal20book_intromanifolds}, we define the retraction on a manifold with corners at $X\in\manifold \hookrightarrow \real^{N}$ as a smooth map $R_{X}: \Tspace_{X}\manifold \rightarrow \manifold$ such that the zero tangent vector maps to $X$, and its differential at the zero tangent vector is the identity map (\ref{eqn:retraction_criteria}).
\begin{align}
\begin{split}
    \retract{X}(0)=&X\\
    \derivative \retract{X}(0)[v] =& v
\end{split}\label{eqn:retraction_criteria}
\end{align} 
We define the Euclidean metric projection (Fig.~\ref{fig:tspace_retraction}c) as a retraction operation, which solves an optimization subproblem~(\ref{eqn:retraction_projection}).
\begin{align}
\begin{split}
    &\retract{X}(v)
    =\argmin\limits_{Y\in\real^{N}} \norm{Y - (X + v)}^2 \\
    &\st h(Y)=0, \;g(Y) \geq 0
\end{split}\label{eqn:retraction_projection}
\end{align}

\subsection{Optimization on CMCs}
\begin{figure}[h]
\vspace{-1em}
\centerline{\includegraphics[width=0.4\textwidth]{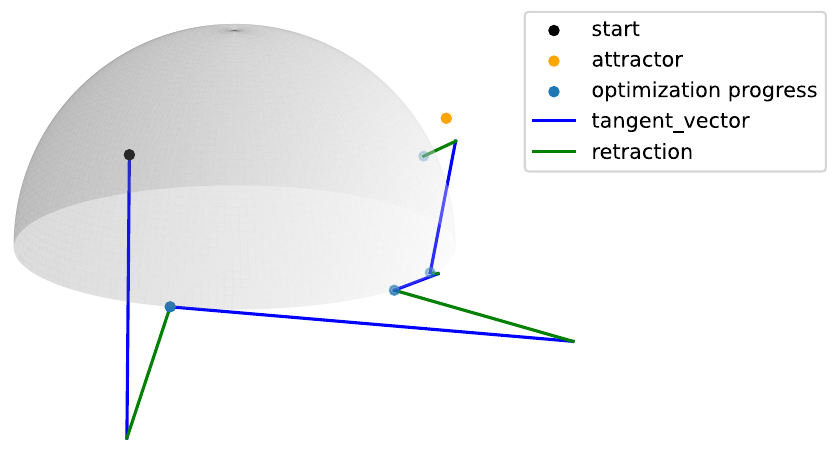}}
\caption{Example of Riemannian gradient descent applied on a ``half sphere" manifold defined by $x^2+y^2+z^2=1$ and $z\geq 0$. We consider a point constrained on the ``half sphere" manifold subject to gravity force pointing downwards and attraction force from a remote point. The optimization minimizes the potential energy of the point.}
\label{fig:half_sphere_traj}
\end{figure}

Finally, we extend Riemannian gradient descent \cite{Boumal20book_intromanifolds} to work on CMCs with the induced metric from the embedding space.
The algorithm (Alg.~\ref{alg:riemannian_gd}) minimizes a real-valued function on CMCs $\mathcal{F}:\mathcal{M}_1\times\ldots\times\mathcal{M}_n\rightarrow \mathbb{R}$, with initial values
$\{X^{(0)}_1\in\manifold_1,\ldots,X^{(0)}_n\in\manifold_n\}$.
In the $k$-th iteration of the algorithm, we first compute the differential of the cost function $\nabla \mathcal{F}(X^{(k)})$, then compute an approximation to the overall gradient by solving the QP subproblem~\eqref{eqn:cmc_gradient} for each manifold. Finally, with a chosen step size $\alpha$, we perform the retraction with \eqref{eqn:retraction_projection} to perform the update and ensures staying on the CMCs.
Fig.~\ref{fig:half_sphere_traj} provides an illustrate toy example.

\vspace{-2em}

\begin{algorithm}[b]
\small
Given $X^{(0)}$, \text{initialize} $k=0$.

\While{not converged}{
a). Compute the differential $\nabla \mathcal{F}(X^{(k)})$ with \eqref{eqn:CMC_differential}.

b). \For{each constraint manifold $\manifold_i$}{
    Compute feasible descent direction $\theta_i^\star$ with \eqref{eqn:cmc_gradient}.
}

c). Choose step size $\alpha$.

d). \For{each constraint manifold $\manifold_i$}{
    Compute tangent vector $v_i^{(k)} = \alpha B_i\theta^*_i$.\\
    Apply retraction $X_i^{(k+1)}=\retract{X^{(k)}_i}(v_i^{(k)})$ with \eqref{eqn:retraction_projection}.
}

e). $k=k+1$
}

\caption{Riemannian gradient descent on CMC}
\label{alg:riemannian_gd}
\end{algorithm}


\section{Experiments and Results}
\subsection{Scenario Setup}
We evaluate the robustness of CMC-Opt on a complex kinodynamic planning problem: a 13-link quadruped (a torso and four 3-link legs) leaping over an obstacle. The trajectory is discretized into 70 time steps and divided into four distinct phases using hybrid collocation \cite{Pardo17RSS_hybrid}: (1) Stance (20 steps) with all four feet grounded, (2) Propulsion (10 steps) where the front legs lift and back legs drive, (3) Flight (20 steps) where all feet are airborne to clear the obstacle, and (4) Landing (20 steps) where all feet land simultaneously and the quadruped returns to its nominal stance pose.

To rigorously test our framework, we adopt the Newton-Euler formulation~\cite{modern-robotics, Xie20arxiv_gtdynamics} for kinodynamics. While highly expressive and general, this approach introduces a massive number of variables and constraints. For every link, the state variables include spatial poses, twists, and accelerations. For every joint, we explicitly model angles, angular velocities, angular accelerations, actuation torques, and transmitted wrenches. Furthermore, we introduce additional variables to explicitly represent contact forces at the feet.

Equality constraints include kinematic constraints (involving link poses, twists, and accelerations) at all joints, alongside Newton-Euler rigid-body dynamics for links.
During ground contact phases, we also include equality constraints to ensure the contact points remain stationary.
Inequality constraints include collision avoidance, Coulomb friction at the contact points, joint angle limits, and actuator torques limits.

Finally, the objective is to minimize a combined weighted sum of squared penalties: control effort (actuator torques), trajectory jerk, state targeting errors (which penalize deviations from desired goal state), and a collocation cost. This final collocation term ensures that the discrete integration of positions and velocities strictly adheres to the trapezoidal rule across all time steps.
To initialize the optimization solvers, we provide a naive, dynamically infeasible initial guess where all twists and accelerations are uniformly set to zero, resulting in a massive collocation violation for the algorithms to overcome.

\subsection{Results}
We benchmark CMC-Opt against four baselines: the penalty method and augmented Lagrangian method are implemented following ~\cite{Vandenberghe20slide_ECNLS}; the trust-region SQP is implemented following ~\cite{Nocedal06book_NumericalOptimization}, and CM-Opt~\cite{Zhang23icra_cmopt} (where inequality constraints are formulated as penalty terms). 
All methods are implemented using the GTSAM library~\cite{Dellaert2012report_gtsam}, employing the Levenberg-Marquardt (LM) algorithm~\cite{Nocedal06book_NumericalOptimization} to solve their respective optimization subproblems.
For CMC-Opt, although Riemannian gradient descent serves as an illustrative framework in earlier sections, the practical implementation deploys a custom LM-like manifold optimization algorithm. This advanced solver leverages Gauss-Newton Hessian approximations and trust-region techniques to ensure robust, rapid convergence.
The performance comparisons are summarized in Table~\ref{tab:benchmark} with four key metrics: state space dimensionality, computational time (in seconds), the constraint violation (in SI units) and cost evaluated at final optimized values.

As Table~\ref{tab:benchmark} shows, CMC-Opt is the only method that drives constraint violations exactly to zero while achieving a significantly lower final objective cost.
By projecting the problem onto the feasible state space, CMC-Opt drastically reduces the search space dimensionality from 32,194 to 2,260. 
As illustrated in Fig.~\ref{fig:quadruped_forward_jump}, CMC-Opt generates a dynamically feasible and natural jumping trajectory. In contrast, the baseline methods fail to yield viable solutions, instead converging to results with extremely high collocation costs.

\begin{figure}[t]
\captionof{table}{Comparison of methods in quadruped planning problem.}
\begin{tabular}{cccccc}
\toprule
 Method &  Dimension &  Time($s$) & Violation & Cost\\
\midrule
Penalty & 32,194 & 1.2e+02 & 1.3e+01  & 2.7e+05 \\
AugL & 32,194 & 5.9e+01  & 7.7e+00 & 3.1e+04 \\
SQP & 32,194 & 8.4e+01 & 1.0e-01 & 1.3e+06 \\
CM-Opt & 2,260 & 8.1e+01 & 4.2e-01 & 7.0e+05 \\
\textbf{CMC-Opt} & \textbf{2,260} & 6.3e+02 & \textbf{0} & \textbf{709.78} \\
\bottomrule
\end{tabular}
\label{tab:benchmark}
\label{tab:cmc_opt_details}
\end{figure}

\section{Discussion}
A noticeable advantage of CMC-Opt over classical constrained optimization methods lies in the significantly reduced problem scale.
This is accomplished through the computation of updates as vectors within the lower-dimensional tangent space of the CMCs.
Furthermore, because the framework internally enforces constraints through CMC retractions, the CMC-Opt approach simplifies the optimization task to a single objective of minimizing the cost function.

Conceptually, the CMC-Opt approach functions as a bi-level optimization framework \cite{Sinha18_bilevel_opt}. The upper-level addresses the core manifold optimization problem, while the lower-level manages manifold-based mathematical operations, including the construction of the tangent space basis and the execution of the retraction operation.
In the trajectory planning example, the variables of the same time step are formulated as a CMC, since the kinodyanmic constraints only connect variables of the same time step.
With this formulation, the higher-level optimization can focus on solving the trajectory optimization problem directly on the lower-dimensional manifolds that reflect the robot's inherent topological structure, while the lower-level optimization is in charge of kinodynamic feasibility.

CMC-Opt shows significant potential for application to robots with highly complex structures.
For systems with contacts or closed-loop kinematics, where defining generalized coordinates is nontrivial, CMC-Opt permits use of the general Euler-Newton formulation, delegating kinodynamic constraints entirely to the lower-level optimization.

CMC-Opt is yet unsuitable for managing cross-time-step constraints, because they cause the constrained-connected component to grow prohibitively large, rendering retraction subproblems intractable to solve.
Such constraints are therefore relaxed as penalty terms in the cost function.

\bibliographystyle{plainnat}
\bibliography{references}

\end{document}